\documentclass{article}
\usepackage[preprint]{neurips_2026}  
\usepackage[utf8]{inputenc}
\usepackage[T1]{fontenc}
\usepackage{hyperref}
\usepackage{url}
\usepackage{booktabs}
\usepackage{amsfonts}
\usepackage{amsmath}
\usepackage{amssymb}
\usepackage{nicefrac}
\usepackage{microtype}
\usepackage{graphicx}
\usepackage{xcolor}

\title{Darkness Visible: Reading the Exception Handler of a Language Model}

\author{
  Peter Balogh \\
  \texttt{palexanderbalogh@gmail.com}
}

\begin{document}
\maketitle

\begin{abstract}
The final MLP of GPT-2 Small exhibits a fully legible routing
program---27 named neurons organized into a three-tier exception handler---while
the knowledge it routes remains entangled across $\sim$3{,}040 residual neurons.
We decompose all 3{,}072 neurons (to numerical precision) into:
5 fused Core neurons that reset vocabulary toward function words,
10 Differentiators that suppress wrong candidates,
5 Specialists that detect structural boundaries, and
7 Consensus neurons that each monitor a distinct linguistic dimension.
The consensus-exception crossover---where MLP intervention shifts from helpful
to harmful---is statistically sharp (bootstrap 95\% CIs exclude zero at
all consensus levels; crossover between 4/7 and 5/7).
Three experiments show that ``knowledge neurons'' \citep{dai2022knowledge},
at L11 of this model, function as routing infrastructure rather than fact
storage: the MLP amplifies or suppresses signals already present in the
residual stream from attention, scaling with contextual constraint.
A garden-path experiment reveals a \emph{reversed} garden-path effect---GPT-2
uses verb subcategorization immediately, consistent with the exception handler
operating at token-level predictability rather than syntactic structure.
This architecture crystallizes only at the terminal layer---in deeper models,
we predict equivalent structure at the final layer, not at layer~11.
Code and data: \url{https://github.com/pbalogh/transparent-gpt2}.
\end{abstract}

\section{Introduction}

The MLP layers of transformer language models are typically treated as opaque
nonlinear transformations.
Recent work has shown that individual neurons can be interpreted
\citep{gurnee2023finding, bricken2023monosemanticity}, that MLP layers implement
key-value memories \citep{geva2021transformer}, and that specific factual
associations can be located and edited \citep{meng2022locating}.
But a complete, legible account of an MLP layer's \emph{routing logic}---readable
as pseudocode with named variables---has remained elusive.

We provide such an account for Layer~11 of GPT-2 Small, the model's final MLP,
while showing that the knowledge it routes remains distributed across
$\sim$3{,}040 residual neurons.
Our decomposition preserves the original weights to numerical precision while
revealing a routing architecture that overturns several assumptions:

\begin{enumerate}
\item \textbf{MLP layers are not opaque.}
A single exception neuron reliably signals which processing path is active,
seven consensus neurons each monitor a distinct linguistic dimension, and
20 exception-handler neurons organize into three functional tiers
(Figure~\ref{fig:pseudocode}).
This routing program is diagnostic, not causal \citep{balogh2026discrete}:
the pseudocode captures the functional organization even though no single
neuron is a causal bottleneck.

\item \textbf{``Knowledge neurons'' are routing infrastructure.}
Neurons identified by \citet{dai2022knowledge} as storing factual knowledge
appear across nearly all facts tested---they are highway signs, not warehouses.
At L11 of GPT-2 Small, this reframes ROME \citep{meng2022locating}: editing
MLP weights changes routing decisions, not stored facts.

\item \textbf{The exception handler has sharp scope boundaries.}
It detects token-level predictability but not syntactic reparse: garden-path
sentences produce a \emph{reversed} effect (\S\ref{sec:garden_path},
Appendix~\ref{app:garden_path}).

\item \textbf{Routing legibility is unique to the terminal layer.}
A survey across all 12 layers shows no comparable structure at any earlier
layer---a phenomenon we call \emph{terminal crystallization}, with a testable
depth-tracking prediction.
\end{enumerate}

\begin{figure}[t]
\centering
\small
\begin{verbatim}
 ROUTING PROGRAM: L11 MLP (27 named neurons)
 -------------------------------------------
 consensus = count(N2, N2361, N2460, N2928,
                   N1831, N1245, N2600)
 exception = N2123.fires    // 11.3% of tokens

 if exception:
   // N2123 is a diagnostic readout, not a causal gate
   CORE(N2123, N2910, N740, N1611, N2044):
     reset vocab -> {the, in, and, a, ,}
     // 54% of output norm, +0.2% PPL
   DIFF(N2462, N2173, N1602, N1800, N2379,
        N1715, N611, N3066, N584, N2378):
     suppress wrong candidates
     repair subword fragments
     // 23% of output norm, +1.3% PPL
   SPEC(N2921, N2709, N971, N2679, N737):
     detect paragraph/clause boundaries
     // 4% of output norm, -0.3% PPL
 else:
   // consensus >= 5/7: MLP intervention
   // is counterproductive (dP < 0)
   // but architecture has no bypass

 RESIDUAL(~3,040 neurons):
   amplify/suppress attention-derived signal
   // distributed, not individually legible
\end{verbatim}
\caption{The exception handler as pseudocode. The pseudocode represents the
\emph{functional organization} observed in activation patterns, not a literal
causal program: N2123 reliably indicates which path is active but does not
control it \citep{balogh2026discrete}.}
\label{fig:pseudocode}
\end{figure}

\begin{quote}
\emph{``No light, but rather darkness visible.''}
\hfill ---Milton, \emph{Paradise Lost}
\end{quote}

\noindent Language models are routinely called ``black boxes''---as if the darkness
inside were uniform.
It is not.
Inside L11's MLP we find \emph{structured} darkness: a legible routing
program wrapped around knowledge that remains opaque.
The darkness is visible precisely because it has structure.

\section{Related Work}

\paragraph{MLP interpretability.}
\citet{geva2021transformer} showed MLPs act as key-value memories;
\citet{geva2023dissecting} traced factual recall through attention and MLP
layers, finding that mid-layer MLPs promote correct attributes---a more nuanced
picture than our binary routing/retrieval framing (\S\ref{sec:knowledge}).
\citet{dai2022knowledge} identified ``knowledge neurons'' in BERT via integrated
gradients; we test their claims on GPT-2 and find routing infrastructure,
not fact storage (\S\ref{sec:knowledge_neurons}).
\citet{bricken2023monosemanticity} decomposed MLP activations via sparse
autoencoders, extracting monosemantic features from superposed representations.
\citet{templeton2024scaling} extended this to production-scale models (Claude 3
Sonnet), demonstrating that SAE decomposition scales and recovers interpretable
features even in large networks.
Our 27 named neurons are legible \emph{without} SAE decomposition---they are
individually interpretable because they serve routing functions.
The $\sim$3{,}040 residual neurons we classify as entangled are precisely the
population where SAE methods should prove most valuable: they encode distributed
knowledge that resists per-neuron interpretation but may yield to
dictionary-learned decomposition.
We view our routing/knowledge partition as complementary to the SAE program:
we identify \emph{which} neurons are entangled and \emph{why} (they encode
content, not control), while SAEs provide the tools to further decompose them.

\paragraph{Circuits and routing.}
\citet{wang2023interpretability} provided a complete circuit for indirect object
identification in GPT-2---tracing a \emph{behavior} across layers; we trace
an \emph{entire layer}'s behaviors within L11's MLP.
\citet{balogh2026discrete} identified the consensus/exception architecture:
7 consensus neurons whose agreement predicts a 94.3pp drop in exception
firing, with the MLP becoming counterproductive at full consensus.
The present paper extends that from \emph{detecting} routing to \emph{reading}
it---characterizing every neuron.
\citet{elhage2021mathematical} introduced the residual stream framework;
\citet{olsson2022context} identified induction heads;
\citet{dettmers2022llm} found sparse outlier features compatible with
binary routing.

\paragraph{Developmental analysis.}
\citet{tenney2019bert} showed BERT rediscovers the NLP pipeline across layers;
\citet{belrose2023eliciting} introduced the tuned lens for layer-wise probing.
Our developmental arc extends these to GPT-2's MLPs with the additional
finding that routing legibility concentrates at the terminal layer.

\section{Methods}

\paragraph{Transparent forward pass.}
We construct a \textsc{TransparentGPT2} wrapping GPT-2 Small (124M parameters)
with no weight modifications.
At Layer~11, the MLP computes $\text{MLP}(x) = W_\text{proj} \cdot h + b_\text{proj}$
where $h = \text{GELU}(W_\text{fc} \cdot x + b_\text{fc})$ is the 3{,}072-dimensional
intermediate activation.
We decompose this into tiers by masking~$h$:
$\text{Tier}_\mathcal{S}(x) = W_\text{proj}[:, \mathcal{S}] \cdot (h \odot m_\mathcal{S})$,
where $\mathcal{S}$ is the neuron index set for each tier and $m_\mathcal{S}$ is
the corresponding binary mask.
The output bias $b_\text{proj}$ is assigned to the Residual tier (it is a
constant offset independent of which neurons fire).
Thus $\text{MLP}(x) = \text{Core}(x) + \text{Diff}(x) + \text{Spec}(x) + \text{Residual}(x)$,
and the sum equals the original MLP output in the 768-dimensional output space
to numerical precision
(max elementwise difference $< 6 \times 10^{-5}$, cosine similarity 0.99999994).

\paragraph{Data.}
All experiments use 512{,}000 tokens from WikiText-103, processed in 500
sequences of 1{,}024 tokens each.
Binary firing: $|\text{GELU}(x_n)| > 0.1$; robustness verified across
$\theta \in \{0.01, 0.05, 0.1, 0.25, 0.5, 1.0\}$ (Appendix~\ref{app:controls}).

\paragraph{N2123 activation regimes.}
At the generic threshold ($|\text{GELU}(x)| > 0.1$), N2123 activates on 70.9\% of tokens.
However, N2123's activation distribution is bimodal: a large low-magnitude
mode centered near 0.3 and a smaller high-magnitude mode above 1.5.
We define the exception-path regime as $\text{GELU}(x_{\text{N2123}}) > 1.0$,
which captures 11.3\% of tokens and corresponds to the upper mode of
this bimodal distribution.
This threshold was selected at the minimum density between the two modes;
results are robust to variation in the range 0.7--1.5 (Appendix~\ref{app:controls}).
Throughout, ``exception path active'' refers to this high-magnitude regime,
while the generic threshold ($> 0.1$) is used only for binary firing patterns
in the enrichment analysis (Table~\ref{tab:enrichment_stats}).

\paragraph{Statistical methods.}
Enrichment uses one-sided Fisher's exact test with Bonferroni correction
($\alpha = 0.05/3072$).
All confidence intervals use sequence-level bootstrap (10{,}000 resamples of
500 sequences, random seed 42) to account for within-sequence autocorrelation.
All statistical tests use \texttt{scipy.stats} (v1.12).

\section{The Exception Handler}
\label{sec:exception}

Among the 11.3\% of tokens where N2123 fires at high magnitude, the remaining
neurons organize into three tiers by conditional fire rate
(Table~\ref{tab:tiers}).

\begin{table}[t]
\centering
\caption{Three-tier exception handler. Fire rates conditional on N2123 firing.}
\label{tab:tiers}
\small
\begin{tabular}{@{}llcccl@{}}
\toprule
Tier & Neurons & Count & Fire Rate & Jaccard & Function \\
\midrule
Core & N2123, N2910, N740, & 5 & 90--100\% & $\geq$0.91 & Vocabulary reset \\
     & N1611, N2044 & & & & \\
Diff & N2462, N2173, N1602, & 10 & 35--88\% & 0.15--0.89 & Candidate suppression, \\
     & N1800, N2379, N1715, & & & & subword repair \\
     & N611, N3066, N584, N2378 & & & & \\
Spec & N2921, N2709, N971, & 5 & 14--37\% & $<$0.15 & Boundary detection \\
     & N2679, N737 & & & & \\
\bottomrule
\end{tabular}
\end{table}

\paragraph{Core: a fused mega-neuron.}
The 5 Core neurons exhibit pairwise Jaccard similarities $\geq 0.91$
(peak: 0.998).
For context, two \emph{independent} neurons each firing at 95\% would produce
Jaccard $\approx 0.90$ by base-rate overlap alone; the observed 0.998 far
exceeds this independence baseline (and the random-init control in
\S\ref{app:controls} yields only 0.53).
Their output directions all push toward function words:
\texttt{the}, \texttt{in}, \texttt{and}, \texttt{a}, \texttt{,}---a
\textbf{vocabulary reset} establishing a generic prior before the residual
contributes contextual adjustments.
The Core accounts for 54\% of exception-path output norm but only $+$0.2\%
PPL when ablated (Table~\ref{tab:ablation})---a DC offset that the residual
stream overwrites.
This is a Simpson's paradox: $+$2.4\% PPL at low consensus (where it matters)
and $-$0.4\% at high consensus (where it doesn't), averaging to the modest
$+$0.2\%.

\paragraph{Differentiators: suppression, not promotion.}
The 10 Differentiator neurons---including a suppression pair
(N584$+$N2378, Jaccard 0.889) and subword repair neurons---contribute 23\%
of output norm and $+$1.3\% PPL.
They suppress wrong candidates rather than promote correct ones, doing the
real discriminative work.

\paragraph{Specialists.}
N737 is a solo paragraph boundary detector (Jaccard $< 0.15$ with all others).
The tier contributes 4\% of output norm and slightly \emph{improves} PPL when
ablated ($-$0.3\%).

\paragraph{The exception indicator.}
N2123 detects tokens where the model ``doesn't yet know what's happening''---it
responds to subword fragments and is inhibited by complete content words.
It is not causal: zeroing it changes PPL by $<$0.1\% \citep{balogh2026discrete}.
It is a \textbf{vote counter}---a readable summary of the distributed routing
decision. Removing the counter does not change the election.

\section{Consensus and the Crossover}
\label{sec:consensus}

\subsection{Seven Dimensions of Normal}

The 7 consensus neurons were identified in \citet{balogh2026discrete}
as neurons that (a)~fire on $>$75\% of tokens overall,
(b)~show $>$10$\times$ enrichment ratio between their most-enriched
and most-depleted token classes (Fisher's exact, Bonferroni-corrected),
and (c)~have output directions with cosine similarity $>$0.4
to the mean MLP output direction at high-consensus positions.
We inherit the same 7 neurons here; independent re-identification
using the same criteria on our 512K-token dataset recovers all 7.
Each monitors a distinct linguistic property
(Table~\ref{tab:consensus_full} in Appendix~\ref{app:consensus}).
Six content neurons (cos 0.52--0.73) form a \textbf{linguistic structure axis};
N2600 (cos $\approx 0.0$ with all others) forms a \textbf{referential
concreteness axis}, firing on currencies, dates, and names while depleting
abstract adjectives.
Full consensus requires both structural predictability \emph{and} referential
concreteness.

\subsection{Consensus Predicts MLP Helpfulness}

The consensus gradient maps directly to whether L11's MLP helps or hurts
(Table~\ref{tab:consensus_help}).

\begin{table}[t]
\centering
\caption{L11 MLP effect on next-token probability by consensus level
(204{,}800 tokens, sequence-level bootstrap; \S3).
All CIs exclude zero; crossover at 4--5/7.}
\label{tab:consensus_help}
\begin{tabular}{@{}crrrl@{}}
\toprule
Level & Tokens & Mean $\Delta P$ & 95\% CI & \\
\midrule
0/7 & 206    & $+$0.187 & [$+$0.145, $+$0.231] & \textcolor{blue}{$\blacktriangle$} \\
1/7 & 913    & $+$0.094 & [$+$0.079, $+$0.109] & \textcolor{blue}{$\blacktriangle$} \\
2/7 & 3{,}051  & $+$0.045 & [$+$0.039, $+$0.051] & \textcolor{blue}{$\blacktriangle$} \\
3/7 & 6{,}283  & $+$0.030 & [$+$0.026, $+$0.033] & \textcolor{blue}{$\blacktriangle$} \\
4/7 & 12{,}617 & $+$0.014 & [$+$0.011, $+$0.016] & \textcolor{blue}{$\blacktriangle$} \\
\midrule
5/7 & 34{,}155 & $-$0.004 & [$-$0.006, $-$0.003] & \textcolor{red}{$\blacktriangledown$} \\
6/7 & 69{,}625 & $-$0.014 & [$-$0.015, $-$0.014] & \textcolor{red}{$\blacktriangledown$} \\
7/7 & 77{,}950 & $-$0.020 & [$-$0.021, $-$0.019] & \textcolor{red}{$\blacktriangledown$} \\
\bottomrule
\end{tabular}
\end{table}

\begin{figure}[t]
\centering
\includegraphics[width=0.85\linewidth]{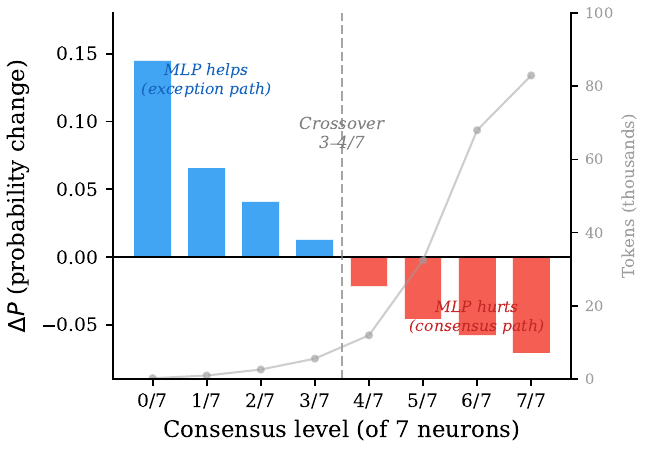}
\caption{L11 MLP effect by consensus level. Blue: MLP helps. Red: MLP hurts.
Crossover between 4/7 and 5/7 matches the causal analysis of
\citet{balogh2026discrete}.}
\label{fig:crossover}
\end{figure}

The crossover between 4/7 and 5/7 is statistically sharp
(Figure~\ref{fig:crossover}): at 4/7 the MLP helps ($\Delta P = +0.014$,
CI $[+0.011, +0.016]$); at 5/7 it harms ($\Delta P = -0.004$,
CI $[-0.006, -0.003]$).
The routing is a \textbf{functional necessity} that correctly identifies when
intervention helps versus harms.

\paragraph{Moloch's compulsion.}%
\footnote{In \emph{Paradise Lost}, Moloch counsels
perpetual war regardless of outcome---``My sentence is for open war''---preferring
action to deliberation. The MLP shares this compulsion: it intervenes on every
token, even when intervention is counterproductive.}
At 7/7 consensus, L11 MLP promotes 7{,}674 tokens to top-1 while losing
3{,}530---a net gain in accuracy but a net \emph{loss} in average probability
($-$0.020). The architecture provides no bypass: every token passes through
GELU, every distribution gets reshaped. The MLP is architecturally incapable
of abstention---a design constraint with implications for efficient inference,
since tokens that attention already predicts correctly still pay the full
computational cost of MLP processing with no benefit.

\subsection{Tier-by-Tier Ablation}

\begin{table}[t]
\centering
\caption{PPL impact of zeroing exception handler tiers (204{,}800 tokens,
sequence-level bootstrap CIs).}
\label{tab:ablation}
\begin{tabular}{@{}lcccc@{}}
\toprule
Tier & Neurons & PPL & $\Delta$ & 95\% CI \\
\midrule
Baseline        & ---  & 31.79 & --- & --- \\
Core            & 5    & 31.86 & $+$0.2\% & [$+$0.1\%, $+$0.4\%] \\
Differentiators & 10   & 32.20 & $+$1.3\% & [$+$0.9\%, $+$1.7\%] \\
Specialists     & 5    & 31.69 & $-$0.3\% & [$-$0.6\%, $-$0.1\%] \\
All exception   & 20   & 32.44 & $+$2.1\% & [$+$1.6\%, $+$2.5\%] \\
\bottomrule
\end{tabular}
\end{table}

\begin{figure}[t]
\centering
\includegraphics[width=0.95\linewidth]{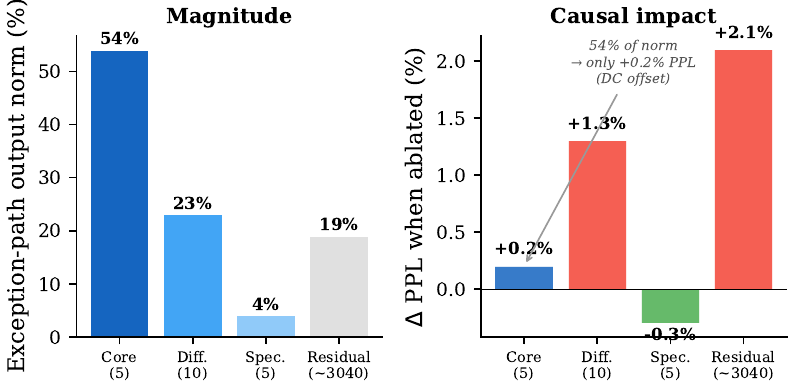}
\caption{The DC offset paradox. \textbf{Left:} Exception-path output norm.
\textbf{Right:} PPL impact. The Core dominates at 54\% of norm but contributes
only $+$0.2\% PPL---a DC offset the residual stream already provides.}
\label{fig:tiers}
\end{figure}

The Differentiators are $6\times$ more important than the Core
(Table~\ref{tab:ablation}), despite contributing less output norm.
Specialists slightly \emph{improve} PPL when ablated---they fire at
unpredictable structural boundaries where their vocabulary push conflicts
with the actual next token.

\section{Where Knowledge Lives}
\label{sec:knowledge}

\subsection{The MLP Does Not Retrieve Facts}

For 160 factual cloze prompts spanning 15 categories, we progressively
accumulate residual neuron contributions to test whether factual knowledge
is stored combinatorially.
Neurons are accumulated in decreasing order of activation-weighted output
norm ($|h_n| \cdot \|W_\text{proj}[:, n]\|_2$), a target-agnostic criterion
that ranks neurons by generic signal magnitude rather than relevance to
any particular answer.
In the \textbf{static} version (raw $W_\text{proj}$ columns), the correct
token never reaches top-10 (0\%).
In the \textbf{context-dependent} version (scaled by actual activations),
only 18/160 (11\%) reach top-10 (median rank 1{,}905).

The category breakdown reveals the mechanism: highly constrained completions
succeed (``300{,}000 km per $\to$ \emph{second},'' rank 1; ``100 degrees $\to$
\emph{C},'' rank 1) while arbitrary associations fail (``France is $\to$
\emph{Paris},'' rank 6{,}092; ``Germany is $\to$ \emph{German},'' rank 45{,}848).
This is primarily \emph{amplification} rather than independent
\emph{retrieval}---the MLP provides contextual constraint satisfaction that
succeeds when the constraint space is small but fails when it is large.
The 11\% success rate for highly constrained completions represents a limited
retrieval capacity, but one dependent on contextual narrowing rather than
stored factual associations.
Full category results appear in Appendix~\ref{app:knowledge}.

\paragraph{Reconciling with mid-layer promotion.}
\citet{geva2023dissecting} found that mid-layer MLPs promote correct attributes
during factual recall---a finding that might seem to contradict our claim that
L11's MLP routes rather than retrieves.
The resolution lies in the developmental gradient across layers:
mid-layer MLPs operate on partially formed predictions where the correct
answer has not yet reached high rank, so \emph{promotion} (boosting the
correct token's logit) is the dominant contribution.
By the terminal layer, attention has already assembled a strong candidate
set in the residual stream.
L11's MLP therefore faces a different computational problem---not ``which
token should be promoted?''\ but ``does the current prediction require
correction?''---which is the routing function we characterize.
This is consistent with terminal crystallization (\S\ref{sec:terminal}):
the routing program we document exists because the terminal layer inherits
a nearly complete prediction and need only intervene when that prediction
is wrong.

\subsection{Knowledge Neurons Are Routing Neurons}
\label{sec:knowledge_neurons}

The neurons \citet{dai2022knowledge} identified as ``knowledge neurons'' are
Belial\footnote{In \emph{Paradise Lost}, Belial ``could make the worse appear /
The better reason''---persuasive but misleading.}---they appear, through the
eloquence of integrated gradients attribution, to store factual knowledge.
They do not. They route.

\paragraph{Attribution overlap.}
Replicating Dai et al.'s method on 20 prompts at L11, on average 7.3 of
each prompt's top-20 attributed neurons are members of our 27-neuron routing
circuit---a 36.5$\times$ enrichment over chance ($27/3072 \times 20 = 0.18$).
The \emph{same} routing neurons appear across nearly all prompts (N2, N611,
N1611, N2044, N2173, N2460, N2600, N2910 each in 16--19 of 20 prompts).
If these neurons stored prompt-specific facts, they should differ for
``Paris,'' ``Berlin,'' ``Jupiter,'' and ``Au.''

\paragraph{Knockout.}
Zeroing Dai's top-20 neurons \emph{increases} target probability by 8.0pp on
average---the opposite of storage.
These neurons push toward function words; removing them lets factual signal
from attention pass through.
Only attention head ablation produces the expected decrease ($-$1.4pp).
Knowledge arrives through the residual stream via attention.

\paragraph{Reconciliation.}
Integrated gradients conflates \emph{influence} with \emph{storage}.
These neurons have high gradients because they are control points: changing
a highway sign has large causal effects without the sign ``storing''
destinations.
Whether Dai et al.'s findings in BERT---a bidirectional model with fundamentally
different information flow---represent genuine storage or also reduce to routing
remains an open question.
Full details in Appendix~\ref{app:knowledge_neurons}.

\section{Terminal Crystallization}
\label{sec:terminal}

If the routing program reflects a general organizational principle of MLPs,
it should appear across multiple layers.
If it is unique to the terminal layer, it reveals something about the model's
developmental trajectory.
We replicated the full characterization at all 12 layers
(Table~\ref{tab:cross_layer}).
The highest Jaccard similarity found at \emph{any} other layer is 0.384 (L10).
L11's Core achieves 0.998---a qualitative gap.
No other layer shows Specialist neurons, suppression pairs, or enrichment
above 1.21$\times$.

\begin{table}[t]
\centering
\caption{Exception handler structure across layers (102{,}400 tokens).
Only L11's known architecture shows tight co-firing.}
\label{tab:cross_layer}
\small
\begin{tabular}{@{}lcccccc@{}}
\toprule
 & L0--L3 & L4--L6 & L7--L9 & L10 & L11 (auto) & L11 (known) \\
\midrule
Exc.\ fire rate & 6--15\% & 12--15\% & 25--36\% & 38.5\% & 10.9\% & 11.3\% \\
Max Jaccard     & 0.06--0.16 & 0.13--0.18 & 0.26--0.36 & 0.384 & 0.123 & \textbf{0.998} \\
Enrichment      & 1.06--1.17 & 1.11--1.21 & 1.06--1.09 & 1.05 & 1.15 & \textbf{2.0} \\
Hi-Jaccard ($>$0.5) & 0 & 0 & 0 & 0 & 0 & \textbf{4} \\
\bottomrule
\end{tabular}
\end{table}

\textbf{Testable prediction:} Legible routing crystallizes at L11 because it is
the \textbf{last opportunity} to adjust predictions.
In deeper models (GPT-2 Medium, 24 layers; Large, 36 layers), equivalent
structure should appear at the final MLP, not at L11.

The logit lens and tuned lens \citep{belrose2023eliciting} confirm a three-phase
developmental arc---scaffold ($\sim$1--2\% top-1 gain/layer), decision
($\sim$5--6\%/layer at L7--L9), terminal---under both probing methods
(Appendix~\ref{app:logit_lens}).

\section{Discussion}
\label{sec:discussion}

\paragraph{The program and the database.}
Our decomposition separates L11's MLP into a legible \textbf{program} (27 named
neurons: binary routing) and an opaque \textbf{database} ($\sim$3{,}040 residual
neurons: contextual adjustment).
The analogy is a software system where the control flow is open-source but the
database is encrypted.

\paragraph{Scope of the exception handler.}
\label{sec:garden_path}
The handler detects token-level vocabulary uncertainty but not syntactic reparse.
A garden-path experiment (15 minimal pairs; Appendix~\ref{app:garden_path})
reveals a \emph{reversed} effect: GPT-2 uses verb subcategorization immediately
(``struggled'' cannot take an object), so the intransitive condition produces
\emph{lower} surprisal at disambiguation than the transitive
($W = 12$, $p = 0.018$).
N2123 shows no differential response ($t(14) = -0.45$, $p = 0.66$;
powered for $d \geq 0.78$).
The exception handler fires on vocabulary uncertainty, not parse ambiguity.

\paragraph{Efficiency.}
Since L11's MLP is counterproductive at high consensus
(Table~\ref{tab:consensus_help}: $\Delta P = -0.020$ at 7/7), bypassing it
for the 40\% of tokens at full consensus would save the MLP forward pass
while \emph{improving} prediction quality on those tokens---a routing-aware
alternative to speculative decoding.
The aggregate PPL cost of such selective bypass is an empirical question we
leave to future work.

\paragraph{Model editing.}
``Knowledge neurons'' are routing infrastructure, reframing ROME
\citep{meng2022locating}: weight edits change routing decisions, not stored
facts. This predicts edits should generalize across phrasings but fail across
domains where routing structure differs.

\paragraph{Limitations.}
Single model (GPT-2 Small), single domain (WikiText-103), limited garden-path
stimuli ($N = 15$), and an underpowered transplant experiment ($N = 5$).
Full discussion in Appendix~\ref{app:limitations}.

\section{Conclusion}

The final MLP of GPT-2 Small exhibits a legible routing program:
7 consensus neurons detect ``normal language'' along distinct linguistic
dimensions, a single exception neuron indicates which of two processing paths
is active, and $\sim$3{,}040 residual neurons provide contextual adjustments
to knowledge already in the residual stream.
This architecture crystallizes only at the terminal layer.

The MLP's contribution is not knowledge retrieval but \textbf{routing}:
deciding whether to intervene or abstain.
We can read 27 neurons as a routing program.
The 3{,}040 neurons they route remain opaque.

\bibliography{references}
\bibliographystyle{plainnat}

\newpage
\appendix

\section{Consensus Neuron Characterization}
\label{app:consensus}

\begin{table}[h]
\centering
\caption{Consensus neuron specializations (512K tokens).}
\label{tab:consensus_full}
\begin{tabular}{@{}clcp{5cm}@{}}
\toprule
Neuron & Dimension & Rate & Key Evidence \\
\midrule
N2     & Clausal continuation   & 88.4\% & Fires on \texttt{and}, \texttt{but}, \texttt{also} mid-clause; silent at clause boundaries \\
N2361  & Syntactic elaboration  & 84.1\% & Fires on \texttt{that}, \texttt{while}, \texttt{neither}, \texttt{fully}; depleted on \texttt{There}, \texttt{United} \\
N2460  & Relational embedding   & 86.0\% & Fires on \texttt{an}, \texttt{into}, \texttt{per}, \texttt{other} in prep.\ phrases; depletes \texttt{According} to 3.3\% \\
N2928  & Sequential structure   & 91.4\% & Fires on ordinals, lists, parallel structure; highest mean activation (0.79) \\
N1831  & Discourse coherence    & 81.0\% & Fires on topic-continuing phrases; 77K disagreement tokens \\
N1245  & Argument structure     & 85.5\% & Fires on \texttt{leadership}, \texttt{validity}, \texttt{one of the}; marks semantic roles \\
\midrule
N2600  & Concrete reference     & 79.2\% & Fires on \texttt{\$}, dates, names; depletes \texttt{natural} (1.8\%), \texttt{social} (3.1\%), abstract adj. \\
\bottomrule
\end{tabular}
\end{table}

The six content neurons (N2--N1245) have pairwise cosine similarities of
0.52--0.73 in output direction space: aligned but not redundant.
N2600 is nearly orthogonal (cos~$\approx 0.0$), forming a separate axis.
Full consensus therefore requires both structural predictability \emph{and}
referential concreteness.

The exception neuron N2123 is aligned with the consensus mean (cosine 0.838):
both routing paths push toward the same safe vocabulary---the difference is
activation context, not direction.

\paragraph{Consensus as linguistic predictability.}
Tokens at 0/7 consensus are 27\% paragraph breaks and rare subwords;
at 7/7 they are dominated by punctuation and function words.
The relationship holds after controlling for frequency: pre-MLP top-1
probability monotonically decreases from 0.249 (7/7) to 0.057 (0/7),
tracking contextual confidence, not token frequency.

\section{Knowledge Extraction: Full Results}
\label{app:knowledge}

\begin{table}[h]
\centering
\caption{Knowledge retrieval across 160 factual prompts.}
\label{tab:twenty_q}
\begin{tabular}{@{}llrrr@{}}
\toprule
Prompt (abbreviated) & Target & Base & Static & Context \\
\midrule
\multicolumn{5}{@{}l}{\emph{Retrievable (context rank $\leq$ 10):}} \\
\ldots 300,000 km per     & second     & 1   & 244    & \textbf{1}     \\
\ldots 100 degrees        & C          & 3   & 68     & \textbf{1}     \\
\ldots created by Linus   & Tor        & 1   & 2{,}551  & \textbf{1}     \\
\ldots London occurred in & 16         & 4   & 372    & \textbf{2}     \\
\ldots begins with four   & notes      & 2   & 3{,}159  & \textbf{2}     \\
\midrule
\multicolumn{5}{@{}l}{\emph{Not retrievable (context rank $>$ 1{,}000):}} \\
\ldots France is          & Paris      & 5   & 3{,}429  & 6{,}092          \\
\ldots language of Germany& German     & 4   & 1{,}402  & 45{,}848         \\
\ldots Russia is          & Moscow     & 7   & 5{,}926  & 33{,}592         \\
\ldots blue               & whale      & 1   & 22{,}915 & 26{,}557         \\
\ldots food from          & Japan      & 2   & 2{,}731  & 48{,}186         \\
\bottomrule
\end{tabular}

\vspace{4pt}
\begin{tabular}{@{}lrrr@{}}
\toprule
Category & $n$ & Top-10 & Median rank \\
\midrule
Historical events  & 10 & 4 & 16 \\
Physics            & 10 & 4 & 93 \\
Music              & 10 & 1 & 554 \\
Mathematics        & 10 & 1 & 594 \\
Technology         & 10 & 3 & 764 \\
Chemistry          & 10 & 1 & 887 \\
Astronomy          & 10 & 1 & 1{,}074 \\
Animals            & 10 & 1 & 1{,}616 \\
Historical people  & 15 & 1 & 1{,}812 \\
Geography (location) & 10 & 0 & 2{,}316 \\
Biology            & 10 & 0 & 3{,}041 \\
Literature/language & 10 & 0 & 3{,}240 \\
Food               & 10 & 0 & 8{,}100 \\
Capitals           & 15 & 0 & 14{,}960 \\
Languages (country) & 10 & 1 & 26{,}530 \\
\midrule
\textbf{All (160)} & \textbf{160} & \textbf{18 (11\%)} & \textbf{1{,}858} \\
\bottomrule
\end{tabular}
\end{table}

\paragraph{Progressive prediction.}
For ``In 1969, astronauts landed on the \underline{\hspace{1cm}}'':
\textbf{Before MLP}: residual predicts \texttt{moon} (from attention).
\textbf{After Core}: shifts toward \texttt{the} (vocabulary reset).
\textbf{After Differentiators}: wrong candidates suppressed.
\textbf{After Residual}: returns to \texttt{moon} with redistributed mass.
Across 12 prompts, 10/12 show this pattern; the 2 exceptions involve
subword-initial predictions.

\section{Knowledge Neurons: Full Details}
\label{app:knowledge_neurons}

\paragraph{Attribution overlap details.}
\begin{table}[h]
\centering
\caption{Integrated gradients overlap with routing neurons (20 prompts).}
\begin{tabular}{@{}lcc@{}}
\toprule
& In top-20 & Expected by chance \\
\midrule
Consensus neurons (7) & 2.6 avg & 0.05 \\
All routing neurons (27) & 7.3 avg & 0.18 \\
Enrichment & 36.5$\times$ & --- \\
\bottomrule
\end{tabular}
\end{table}

The 36.5$\times$ enrichment uses a uniform baseline ($27/3072 \times 20 = 0.18$).
Since top-20 integrated-gradient neurons are selected for high causal influence,
and routing neurons have high influence by definition, the true enrichment over
a causally-matched baseline would be lower.
The qualitative finding---that the \emph{same} routing neurons recur across
diverse facts---is the more robust evidence.

\paragraph{Knockout details.}
Of the 20 prompts, 17 showed increased target probability when Dai's neurons
were zeroed (range: $+$0.3pp to $+$24.3pp), 2 showed negligible change
($< \pm$0.1pp), and 1 showed a decrease ($-$2.1pp).
The consistency across diverse facts (85\% showing improvement) confirms
that these neurons systematically suppress factual signal rather than
occasionally doing so.
The apparent tension between knockout improving factual recall ($+$8.0pp) while
full exception ablation worsens PPL ($+$2.1\%) resolves because the handler
is optimized for the common case (function-word prediction at low consensus)
at the cost of rare factual completions.

\paragraph{Activation transplant.}
For 5 country-capital pairs, transplanting Dai's top-20 neuron activations from
source to destination prompt transplanted no facts (0/5).
Source-fact probability changed $<$0.02pp; this experiment ($N = 5$) is
illustrative, not standalone evidence.

\section{Logit Lens and Developmental Arc}
\label{app:logit_lens}

\begin{table}[h]
\centering
\caption{Top-1 accuracy via logit and tuned lens \citep{belrose2023eliciting}
(204{,}800 tokens). ``First lock-in (Tuned)'' = percentage of tokens that
first reach top-1 accuracy at this layer under the tuned lens.}
\label{tab:logit_lens}
\begin{tabular}{@{}lcccc@{}}
\toprule
Layer & Logit Top-1 & Tuned Top-1 & $\Delta$ & First lock-in (Tuned) \\
\midrule
Emb    & 0.7\%  & 3.5\%  & $+$2.8  & 3.5\% \\
L0     & 3.0\%  & 9.0\%  & $+$6.0  & 6.2\% \\
L1     & 3.9\%  & 9.7\%  & $+$5.8  & 1.2\% \\
L2     & 3.9\%  & 10.7\% & $+$6.7  & 1.4\% \\
L3     & 5.0\%  & 11.7\% & $+$6.6  & 1.4\% \\
L4     & 5.9\%  & 12.6\% & $+$6.7  & 1.3\% \\
L5     & 8.6\%  & 16.2\% & $+$7.6  & 4.0\% \\
L6     & 10.8\% & 19.0\% & $+$8.2  & 3.4\% \\
\midrule
L7     & 17.0\% & 24.6\% & $+$7.6  & 5.9\% \\
L8     & 22.7\% & 29.7\% & $+$6.9  & 6.0\% \\
L9     & 29.5\% & 34.9\% & $+$5.4  & 5.9\% \\
\midrule
L10    & 34.4\% & 37.3\% & $+$2.9  & 3.7\% \\
L11    & 39.0\% & 39.0\% & $+$0.0  & 3.0\% \\
\midrule
Never  & ---    & ---    & ---     & 53.0\% \\
\bottomrule
\end{tabular}
\end{table}

\begin{figure}[h]
\centering
\includegraphics[width=0.85\linewidth]{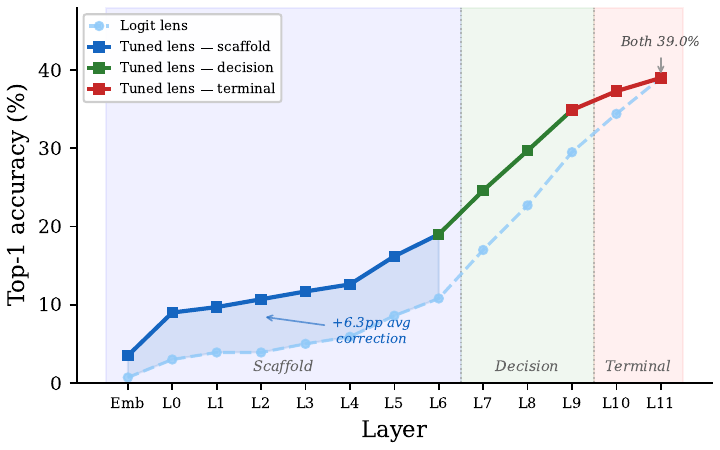}
\caption{Top-1 accuracy across layers. The developmental arc holds under both
logit lens (dashed) and tuned lens (solid).}
\label{fig:logit_lens}
\end{figure}

The tuned lens corrects early-layer underestimates (mean $+$6.3pp at L0--L3)
while converging at L11 ($+$0.0pp).
The developmental arc is confirmed: decision-phase layers gain 5.3pp/layer
under tuned lens vs.\ 1.5pp in scaffold phase.

\paragraph{Illustrative cases.}
\textbf{Easy:} ``Abraham $\to$ Lincoln'' locks in at L0; L11 MLP slightly
reduces confidence.
\textbf{Medium:} ``moon'' overtakes ``planet'' at L10's MLP.
\textbf{Hard:} ``second'' (speed of light) reaches top-1 only at L11.

\paragraph{L11H7: the dominant attention head.}
L11's most important head (6$\times$ the next) sends 45.4\% of weight to BOS---the
attention sink phenomenon \citep{xiao2023efficient}. Exception tokens attend to
BOS even more (47.0\% vs 37.3\%).

\section{Garden-Path Experiment}
\label{app:garden_path}

Following \citet{vangompel2001lexical}, we constructed 15 minimal pairs
(Table~\ref{tab:garden_path_full}).
In each pair, the intransitive verb cannot take a direct object, forcing
the post-verbal NP to be parsed as a new clause subject.
The transitive verb is ambiguous: the post-verbal NP could be its object.

\begin{table}[h]
\centering
\caption{All 15 garden-path minimal pairs with surprisal at disambiguation.
Pairs 1--5 follow \citet{vangompel2001lexical}; pairs 6--15 extend the set.}
\label{tab:garden_path_full}
\small
\begin{tabular}{@{}clllrr@{}}
\toprule
\# & Intrans.\ verb & Trans.\ verb & Disambig. & $S_\text{int}$ & $S_\text{trans}$ \\
\midrule
1  & struggled  & scratched  & took      & 5.4  & 11.6 \\
2  & sneezed    & visited    & prescribed & 6.8  & 16.2 \\
3  & dozed      & watched    & next      & 8.9  & 11.5 \\
4  & escaped    & attacked   & searched  & 17.0 & 19.4 \\
5  & slept      & ignored    & standing  & 13.2 & 11.3 \\
6  & cried      & woke       & in        & 9.0  & 6.9  \\
7  & purred     & bit        & sitting   & 14.5 & 14.2 \\
8  & fainted    & alarmed    & on        & 9.3  & 7.5  \\
9  & galloped   & threw      & on        & 5.4  & 5.7  \\
10 & erupted    & destroyed  & of        & 1.6  & 1.3  \\
11 & sputtered  & startled   & under     & 12.8 & 11.9 \\
12 & lied       & accused    & at        & 6.9  & 8.2  \\
13 & performed  & impressed  & in        & 5.0  & 5.3  \\
14 & marched    & disobeyed  & at        & 7.4  & 8.5  \\
15 & blazed     & trapped    & near      & 12.2 & 8.3  \\
\midrule
\multicolumn{4}{@{}l}{Median $\Delta$ (trans $-$ int)} & \multicolumn{2}{c}{$+$3.1 bits} \\
\bottomrule
\end{tabular}
\end{table}

\paragraph{Example.}
\begin{quote}
\emph{After the dog \textbf{struggled} the vet took off the muzzle.} \\
\emph{After the dog \textbf{scratched} the vet took off the muzzle.}
\end{quote}

In the intransitive version, human readers initially parse ``the vet'' as
the object of ``struggled,'' then reparse at ``took.''
We measured surprisal, consensus, N2123, and MLP delta at disambiguation.

\paragraph{Results.}
No differential N2123 response (0.102 vs 0.105, $t(14) = -0.45$, $p = 0.66$;
powered for $d \geq 0.78$).
The transitive condition produces \emph{higher} surprisal at disambiguation
in 10 of 15 pairs (Wilcoxon $W = 12$, $p = 0.018$, median $+$3.1 bits).
GPT-2 uses verb subcategorization immediately: ``struggled'' cannot take an
object, so ``the vet'' is parsed as a new-clause subject from the start.
The garden-path is \emph{reversed}: transitive verbs create the reparse.

\paragraph{Connection to selective modularity.}
\citet{britt1992parsing} showed that discourse context can override shallow
attachment preferences but not deep clause-level reparse. Our verb
subcategorization ambiguity is structurally analogous: GPT-2 resolves it
immediately without exception-handler involvement, paralleling Britt et al.'s
autonomous syntactic component.

\section{Controls}
\label{app:controls}

\paragraph{Null model.}
A randomly initialized GPT-2 (same architecture, untrained weights) shows
no structure:

\begin{center}
\small
\begin{tabular}{@{}lcc@{}}
\toprule
Metric & Trained & Random init \\
\midrule
Exception fire rate & 11.3\% & $\sim$64\% \\
Core max Jaccard    & 0.998  & 0.53 \\
Consensus--exception anticorrelation & 94.3pp range & $<$2pp (flat) \\
Enrichment (max)    & 2.0$\times$ & 1.02$\times$ \\
\bottomrule
\end{tabular}
\end{center}

\paragraph{Threshold robustness.}
Core co-firing (Jaccard $\geq 0.45$) and consensus-exception anticorrelation
(spread $\geq 0.11$) persist across $\theta \in \{0.01, 0.05, 0.1, 0.25, 0.5, 1.0\}$.
No threshold produces qualitatively different structure.

\begin{table}[h]
\centering
\caption{Enrichment of routing neurons under exception firing
(Fisher's exact, Bonferroni-corrected).}
\label{tab:enrichment_stats}
\begin{tabular}{@{}llrrrr@{}}
\toprule
Neuron & Tier & Base rate & Exc rate & Enrichment & $p$-value \\
\midrule
N2123 & Core           & 70.9\% & 100.0\% & 1.41$\times$ & $< 10^{-300}$ \\
N584  & Differentiator & 6.2\%  & 8.3\%   & 1.33$\times$ & $< 10^{-300}$ \\
N2378 & Differentiator & 5.7\%  & 7.4\%   & 1.30$\times$ & $< 10^{-300}$ \\
N737  & Specialist     & 4.0\%  & 4.9\%   & 1.22$\times$ & $< 10^{-300}$ \\
N1602 & Differentiator & 22.7\% & 24.4\%  & 1.08$\times$ & $< 10^{-300}$ \\
N611  & Differentiator & 67.2\% & 71.8\%  & 1.07$\times$ & $< 10^{-300}$ \\
\midrule
\multicolumn{2}{@{}l}{Consensus (5/7)} & \multicolumn{2}{c}{---} & 0.96--1.01 & $> 0.05$ \\
\multicolumn{2}{@{}l}{Remaining (15)} & \multicolumn{2}{c}{---} & 0.98--1.01 & $> 0.01$ \\
\bottomrule
\end{tabular}
\end{table}

\section{Limitations}
\label{app:limitations}

\begin{itemize}
\item \textbf{Single model}: All analysis is on GPT-2 Small (124M). Terminal
crystallization in deeper models is predicted but untested.
\item \textbf{Single domain}: WikiText-103 (encyclopedic prose). Consensus
characterizations may not hold for dialogue, code, poetry, or legal text.
\item \textbf{Single layer depth}: L11 is characterized comprehensively; L7 and
L10 only sketched.
\item \textbf{Garden-path stimuli}: 15 pairs testing verb subcategorization only;
reduced relatives may differ.
\item \textbf{BPE confounds}: Disambiguation words may fall at different absolute
positions due to tokenization.
\item \textbf{Transplant}: $N = 5$; treated as consistency check only.
\item \textbf{SAE alternative}: We characterize neurons individually rather than
via sparse autoencoders \citep{bricken2023monosemanticity, templeton2024scaling},
which might resolve additional structure within the $\sim$3{,}040 residual
neurons our manual method classifies as entangled.
\end{itemize}

\end{document}